\documentclass[conference]{IEEEtran}
\IEEEoverridecommandlockouts
\usepackage{cite}
\usepackage{amsmath,amssymb,amsfonts}
\usepackage{algorithmic}
\usepackage{graphicx}
\usepackage{textcomp}
\usepackage{xcolor}
\usepackage{latexsym}
\usepackage{hyperref}
\usepackage{amssymb}
\usepackage{amsmath}
\usepackage{amsthm}
\usepackage{booktabs}
\usepackage{enumitem}
\usepackage{graphicx}
\usepackage{color}
\usepackage{array}
\usepackage{multicol}
\usepackage{multirow}
\usepackage{wrapfig}
\usepackage{stfloats}
\usepackage{wrapfig}
\usepackage{float}
\usepackage{todonotes}
\usepackage{bbm}

\def\BibTeX{{\rm B\kern-.05em{\sc i\kern-.025em b}\kern-.08em
    T\kern-.1667em\lower.7ex\hbox{E}\kern-.125emX}}
\begin{document}

\title{Session-based Recommender Systems: User Interest as a Stochastic Process in the Latent Space
\thanks{This work was supported by the Polish National Science Centre (NCN) under grant OPUS-18 no. 2019/35/B/ST6/04379. Calculations have been carried out using resources provided by Wroclaw Centre for Networking and Supercomputing, Grant No. 405.}
}

\author{\IEEEauthorblockN{1\textsuperscript{st} Klaudia Balcer}
\IEEEauthorblockA{\textit{Computational Intelligence Research Group} \\
\textit{Institute of Computer Science}\\
\textit{University of Wroclaw}\\
0009-0001-4176-087X \\
klaudia.balcer@cs.uni.wroc.pl}
\and
\IEEEauthorblockN{2\textsuperscript{nd} Piotr Lipinski}
\IEEEauthorblockA{\textit{Computational Intelligence Research Group} \\
\textit{Institute of Computer Science}\\
\textit{University of Wroclaw}\\
0000-0002-6928-262X \\
piotr.lipinski@cs.uni.wroc.p}
}




\maketitle

\begin{abstract}
This paper jointly addresses the problem of data uncertainty, popularity bias, and exposure bias in session-based recommender systems. We study the symptoms of this bias both in item embeddings and in recommendations. We propose treating user interest as a stochastic process in the latent space and providing a model-agnostic implementation of this mathematical concept. The proposed stochastic component consists of elements: debiasing item embeddings with regularization for embedding uniformity, modeling dense user interest from session prefixes, and introducing fake targets in the data to simulate extended exposure.
We conducted computational experiments on two popular benchmark datasets, Diginetica and YooChoose 1/64, as well as several modifications of the YooChoose dataset with different ratios of popular items. The results show that the proposed approach allows us to mitigate the challenges mentioned. Our code is available at \href{https://github.com/presented-after-review-period}{https://github.com/presented-after-review-period}.
\end{abstract}


\section{Introduction}
Recommender Systems (RS) are artificial intelligence models that are used to serve personalized content based on users' previous actions. The characteristic of the data collected for model training vary depending on the environment of user activity. When the user is asked to rate an item directly, for example a~purchased product, we gather explicit feedback \cite{Huang_2024{"originatingScript":"m2","payload":{"guid":"42af076b-f9af-4016-8f3a-5fa785a134d25ada3b","muid":"036f06dc-1d4f-450c-95e4-b81319765d50381910","sid":"457f3189-81ad-4d9c-a513-a4a62d5053d0d33010"}}{"originatingScript":"m2","payload":{"guid":"42af076b-f9af-4016-8f3a-5fa785a134d25ada3b","muid":"036f06dc-1d4f-450c-95e4-b81319765d50381910","sid":"c866a958-496e-445e-a47a-101bbca1df1d4a6457"}}}. If the user has the ability to express positive or negative feedback to content on streaming platforms or social media, we gather explicit binary feedback \cite{Wu_2021}. On the other hand, when following user activity without asking for direct expression of interest, we gather implicit feedback. In such a situation, we assume that the user action is an expression of interest. It can also be measured both binary (was there an action like reading a piece of news) \cite{Wu2019ke} and numerically (how long the action was, for example, the time spent playing a game) \cite{Joachims_2017_implicit}. The purpose of the training procedure is to recover the user activities excluded from the train data. Thus, we build a model that proposes items according to the user's interests, judged from their previous behavior. Regardless of the exact scenario, most modern state-of-the-art RS are based on deep neural networks.

In our research, we focus on Session-Based Recommender Systems (SBRS). In this scenario, we consider anonymous sessions consisting of user interactions (for example, clicks) and provide next-item recommendations \cite{Wu2019ke}. The task is to predict user interest (the next click) given only the session prefix.
Building SBRS comes with several challenges, such as mining implicit feedback and extreme data sparsity, among others \cite{10.1145/3465401}. The biases present in the data are then further propagated into the model and its output. In this work, we focus on the uncertainty of the data, the popularity bias, and the exposure bias. Uncertainty is related to non-deterministic user behavior. Popularity bias is the influence of item popularity on users' behavior, and thus the session data. Exposure bias is caused by the limitations of the visibility of the items. 
The issues described above are closely related and therefore should be addressed jointly. 

We propose treating sessions as a stochastic process in the latent space and its model-agnostic implementation as a stochastic component to be added to a deep model. We conduct computational experiments using two popular benchmark datasets Diginetica and YooChoose 1/64 and several modifications of the YooChoose dataset with different ratios of popular items. We evaluate the proposed approach in terms of recommendation quality and symptoms of bias identified in the preliminary study, including recommendation accuracy, popularity, coverage, and embedding distribution. 

Our contribution can be summarized as follows:
\begin{itemize}
    \item we identify symptom of popularity bias in SBRS, it is encoding popularity in item embeddings distribution;
    \item we formulate sessions as stochastic process in the latent space to jointly address data uncertainty, popularity and exposure bias;
    \item we provide an model-agnostic implementation of the mathematical idea consisting of three components: debiasing item embeddings, modeling dense user interest, extending user exposure by introducing \textit{fake targets};
    \item we conduct computational experiments and broadly evaluate our approach on datasets of varying levels of bias.
\end{itemize}


\section{Session-based Recommender Systems} \label{sec:background}

\subsection{Formal Task Statement}

In this paper, we consider the next-item prediction task for anonymous sessions. Formally, let us consider a set of $N$ items $\mathcal I = \{i_1, i_2, \ldots, i_N\}$. Historical data is a set of $M$ anonymous sessions $\mathcal S = \{S_1, S_2, \ldots, S_M\}$, where each session is a list of items $S_m = (s_{m,1}, s_{m,2}, \ldots, s_{m,L_m - 1})$ of length $L_m$ (where each $s_{m, l} \in \mathcal I$). Given the session prefix $S_{m,{1:L_m-1}} = (s_{m,1}, s_{m,2}, \ldots, s_{m,L_m-1})$, our task is to provide $K$ recommendations $R_{m} = (r_{m,1}, r_{m,2}, \ldots, r_{m,K})$, preferably including the target item $s_{m,L_m}$: 
$S_{m,{1:L_m-1}} \xrightarrow{RS}R_m, \text{ preferably } s_{m,L_m} \in R_m$.

SBRS can be evaluated online and offline. The online evaluation is based on real user feedback, while in the offline evaluation, we check if the recommendations mimic historical user interest stored in the data. In this paper, we use the off-line evaluation scenario. 

\subsection{Latent Item Representations} \label{sec:emb}

Latent representations, called embeddings, are $d$-dimensional vectors optimized during the learning of neural networks in modeling multiple data modalities to represent the object knowledge available in the data. They often provide us with additional insights that are difficult to extract using conventional methods. Embeddings allow us to extract features from unstructured data, such as images, as well as to represent complex relationships between discrete tokens in natural language modeling, among other applications. The importance of good-quality representations led to the separation of representation learning as a branch of deep learning. 

Most recommender systems are based on latent item representations \cite{Wu2019ke, He2017}. 
As we base on purely collaborative signals, the similarities encoded in the latent space are based on the context the item is occurring in within the sessions. Let us denote the embedding set as $E = \{e_1, e_2, \ldots, e_N\}$, where $e_n$ is the embedding vector of the item $i_n$ and $e_{m, j}$ is the embedding vector of the item $s_{m, j}$. 

The inner workings of an SBRS can be described on a very high level as follows. We provide the latent item representations of all items $E$. We process each session prefix $S_{m, L_m-1}$ using item embeddings $(e_{m, 1}, e_{m, 2}, \ldots, e_{m, L_m -1})$ to obtain a vector representing the predicted user interest $\mathbf{s}_m$. Then we score each item based on the similarity between its embedding $e_n$ and the vector $\mathbf{s}_m$. In the basic scenario, we recommend the $K$ items with the highest scores. 
The similarity score between two items is the dot product of their embeddings. 

\subsection{Open Challenges}

Developing and using SBRS is associated with several challenges, such as data sparsity (each session contains only a tiny part of the available items) and dynamic item set (the items may be added and removed from the environment), among others. In this work, w focus on three tightly related challenges: data uncertainty, popularity and exposure bias.

\textbf{Data uncertainty} is intrinsically related to the observation of human behavior \cite{Coscrato_2023}. Human actions, including clicks and purchases in an e-commerce, are influenced by multiple internal and external factors, such as mood, context, season and marketing campaigns. Additional complexity of the problem is introduced by implicit feedback, where we assume that interactions is equivalent to satisfaction, which is not necessarily true.  

\textbf{Popularity bias} is a tendency to favor a small number of the most popular items over the less frequent ones \cite{Klimashevskaia_2024, Huang_2024}. The bias is present in most datasets and then further propagated into the model and recommendations. As many items are underrepresented in the historical data, we do not provide high-quality latent representations of those, and they are even less frequent in the recommendations. 

\textbf{Exposure bias} is the influence of item visibility on user choices (interactions made) \cite{Krause_2024}. In session-based recommender systems, where interactions such as clicks are both cheap and rapid, this effect is particularly strong. Selecting only from a small number of displayed items means we prefer an item over the rest of served ones (but not necessarily over all available ones what we assume when not having information about the exposure).

Those three challenges vary by the measurability; while popularity is easy to measure, it is hard to estimate uncertainty of each user action and directly evaluate the impact of exposure using only interaction data. However, they are tightly interconnected. The user is often exposed to the most popular items, directly on the platform and by external advertisement, and the biases influence user actions. Thus, we address them jointly.

\section{Related Work} \label{sec:related}





\subsection{Literature Overview}



SBRS modeling methods are constantly evolving with advances in the field of deep learning. 
Early work on SBRS was based on Markov chains \cite{Ji_2015}. Then adaptations of recurrent neural networks have been used \cite{improved_rnn_for_session_based_recommendations}. A Gated Recurrent Unit for RS was proposed in GRU4Rec \cite{hidasi2016sessionbased}. A recent breakthrough in session-based recommendations has been achieved using GNNs, as in SR-GNN \cite{Wu2019ke}. This approach has been further enhanced using mechanisms with attention in TAGNN \cite{Yu_2020} and self-attention, as in TAGNN++ \cite{tagnnpp} and STAR \cite{Yeganegi_2024}, highway networks, as in SGNN-HN \cite{sgnnhn}, and improved predictor modules, as in SR-PredAO+DIDN \cite{srgnn-predicton}, among others.
GNNs have also been used as a basis for contrastive learning solutions, such as RESTC \cite{10.1145/3626091}. It is worth mentioning that in the case of RS the data required for supervised and unsupervised learning remain the same.


Exposure bias is a challenge present in different types of RS, including causality-based \cite{LiuUnknown}, sequential \cite{Yang_2024}, and session-based \cite{faster} RS. There are multiple approaches to dealing with the problem, including exposure estimation \cite{Zhou_2021}, self-supervised hard negative mining \cite{Wu_2021} and post-processing techniques \cite{faster}.  However, these approaches are not necessarily easily transferable between different scenarios. 
Similarly, it is for popularity bias, which can be addressed with data augmentation \cite{Wu_2021}, aggregation during training \cite{Zhou_2023} or representation normalization \cite{Gupta2019NISERNI}, among others. There are also several approaches to dealing with popularity together \cite{Abdollahpouri_2017, Zhou_2023} and coverage with popularity together \cite{Liu_2023}.

In modeling data uncertainty, we are not provided with any sort of ground truth. Instead, we try to estimate how certain we are about the data we are using to develop the model. Various attempts are proposed to address this issue, starting from using additional diverse sources of knowledge \cite{Xiong_2024} to building stochastic representations instead of classical deterministic ones \cite{Jiang2020}. Some RS estimate uncertainty simultaneously with learning user preferences \cite{10177891} or provide appropriate selection policies \cite{Wang_2023}.

Despite diverse approaches to modeling uncertainty, popularity, and exposure bias, these problems are seldom addressed together and the work for SBRS remains limited. 




\subsection{SR-GNN}


In this paper, we focus mainly on the GNN-based approaches. The starting point for our approach is the basic SR-GNN. In graph-based RS, each session 
$S_m$ is encoded in the form of a session graph, being a directed graph $\mathcal{G}_m$,
where the set of nodes $\mathcal{V}_S = \{v_1, v_2, \ldots, v_{L_m}\}$ contains the nodes $v_1, v_2, \ldots, v_{L_m}$ corresponding to items $s_{m,1}, s_{m,2}, \ldots, s_{m,L_m}$, respectively, and the set of edges $\mathcal{E}_S$ contains edges $(v_k, v_j)$ such that the user clicked $i_j$ directly after $i_k$ in the session $S_m$. SR-GNN starts with constructing a latent vector embedding for each item (an embedding layer), so it transforms each vertex $v_i$ into a product embedding vector $e_i \in \mathbb{R}^d$ in the latent embedding space $\mathbb{R}^d$. Afterwards, it processes the entire session, as a sequence of the embeddings of the items, through a Gated Graph Neural Network (GGNN), according to the following equations:
\begin{align}
    \mathbf{a}^{(t)}_{s,i} &= \mathbf{A}_{s,i}[e_1^{(t-1)}, e_2^{(t-1)}, \ldots, e_n^{(t-1)}]^\top \mathbf{H} + \mathbf{b}, \\
    \mathbf{z}^{(t)}_{s,i} &= \sigma \left( \mathbf{W}_z \mathbf{a}^{(t)}_{s,i} + \mathbf{U}_z e^{(t-1)}_i \right) \\
    \mathbf{r}^{(t)}_{s,i} &= \sigma \left( \mathbf{W}_r \mathbf{a}^{(t)}_{s,i} + \mathbf{U}_r e^{(t-1)}_i \right) \\
    \tilde{e}^{(t)}_i &= \tanh \left( \mathbf{W}_o \mathbf{a}^{(t)}_{s,i} + \mathbf{U}_o \left( \mathbf{r}^{(t)}_{s,i} \odot e^{(t-1)}_i \right) \right) \\
    e_i^{(t)} &= \left( 1 - \mathbf{z}^{(t)}_{s,i} \right) \odot e^{(t-1)}_i + \mathbf{z}^{(t)}_{s,i} \odot e^{(t)}_i,
\end{align}

\noindent where $\mathbf{z}_{s,i}$ is the reset gate, $\mathbf{r}_{s,i}$ is the update gate, $\tilde{e}_{s,i}$  is the output, $\mathbf{A}_s \in \mathbb{R}^{d \times 2d}$ is the adjacency matrix of the session graph (incoming and outgoing edges concatenated), $\mathbf{A}_{s,i}$ is its $i$-th row corresponding to the node $v_{s,i}$, $\mathbf{H} \in \mathbb{R}^{d \times 2d}$ and $\mathbf{b} \in \mathbb{R}^d$ are the weight matrix and the bias vector, respectively. $\mathbf{W}_z, \mathbf{U}_z, \mathbf{W}_r, \mathbf{U}_r, \mathbf{W}_o, \mathbf{U}_o \in \mathbb{R}^{d \times d}$ are the weights matrices, $\sigma$ is the sigmoid function, and $\odot$ denotes the element-wise multiplication.

\noindent After processing the sequence of nodes, it transforms the entire session into a latent session embedding (linear layers). Finally, SR-GNN predicts recommendations by computing recommendation score $z_j$ for each product $i_j \in \mathcal{I}$ by evaluating the dot product between the session embedding $\mathbf{s}$ and the product embedding and transforming it by a softmax function: $\hat{y}_j = \mbox{softmax}(\mathbf{s}^T \cdot e_j)$,
where $\hat{y}_j$ is the probability of the product $i_j$ being the next product to be browsed by the user. The model is optimized with the binary cross entropy loss function:
\begin{equation}
    \mathcal L = \sum_{j=1}^N y_j \cdot \log(\hat y_j) + (1 - y_j) \cdot \log(1 - \hat y_j)
\end{equation}

\section{Symptoms of Bias in SBRS} \label{sec:preliminaries}




In the preliminary study, we focused on the symptoms of bias encoded in the model itself and its output. The goal was to assess how uncertainty, popularity bias, and exposure bias influence the model in terms of latent item representation and recommendations. We conducted computational experiments on the YooChoose 1/64 dataset and the SR-GNN \cite{Wu2019ke} model. We discuss symptoms known from other papers as well as identify another one related to embedding distribution. 

\begin{wrapfigure}{r}{0.4\linewidth}
    \centering
    \includegraphics[width=1\linewidth]{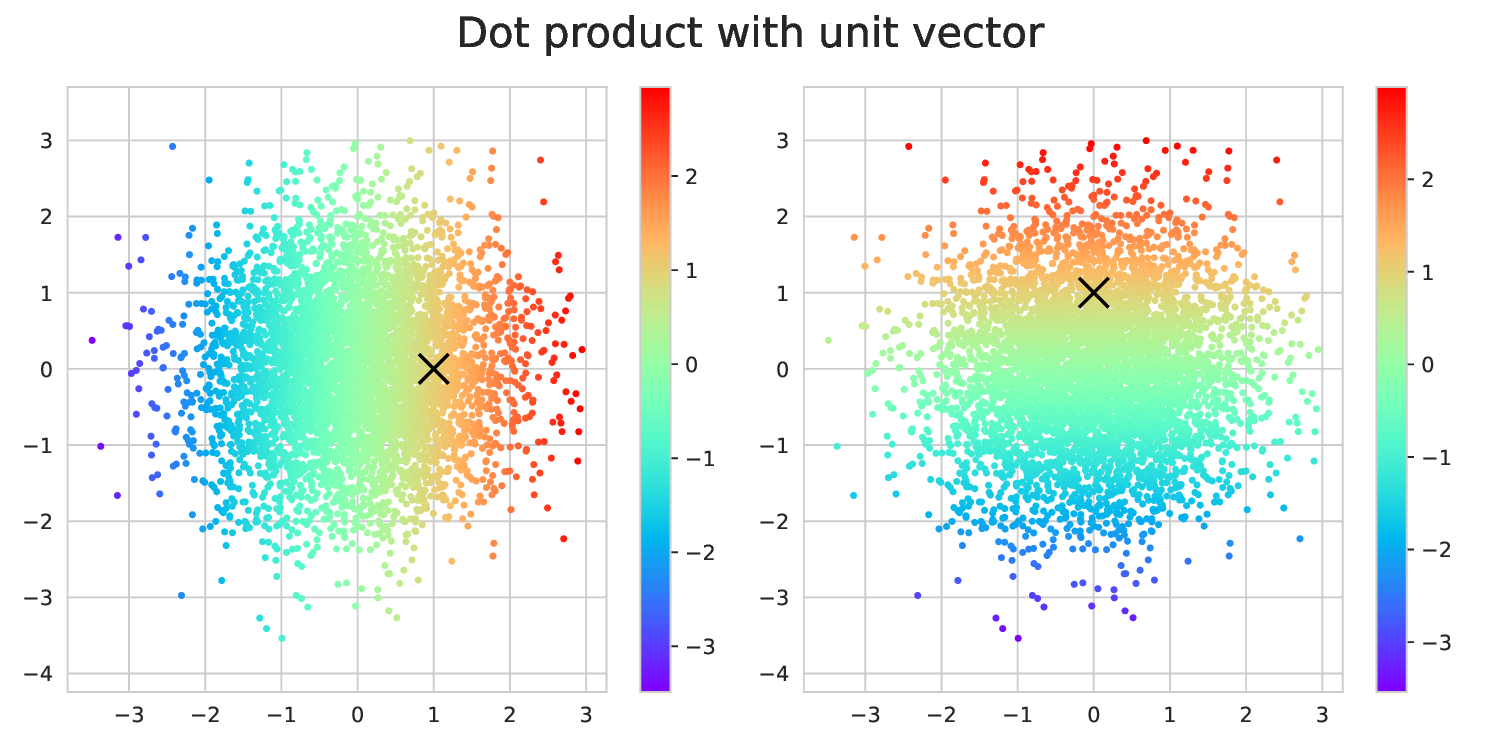}
    \caption{Points represent 2D random embeddings. Points marked with 'x' (one per plot) represent the predicted session embeddings. Dot-products between the session embeddings and the item embeddings have been calculated and marked with the colour. Products closest to red would be recommended
    }
    \label{fig:dot-product}
\end{wrapfigure}

First, let us start with an illustration of how the norms of the embedding vectors can influence the recommendations. Figure~\ref{fig:dot-product} presents example 2D item embeddings (each point corresponds to one item) and predicted user interest (marked with 'x'). The colors present scores of items calculated with the dot product, which depend on both the cosine similarity and the embedding norm. We can see that the models favor items with longer embeddings. Encoding popularity in embedding norm has been already referred in the research \cite{Gupta2019NISERNI}. 

\begin{wrapfigure}{r}{0.4\linewidth}
    \centering    \includegraphics[width=\linewidth]{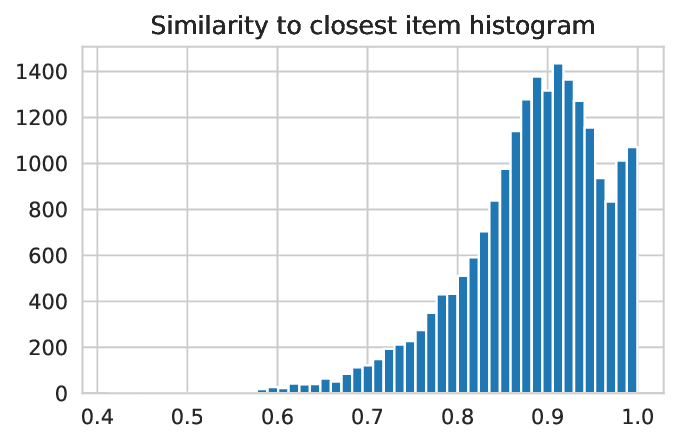}
    \includegraphics[width=\linewidth]{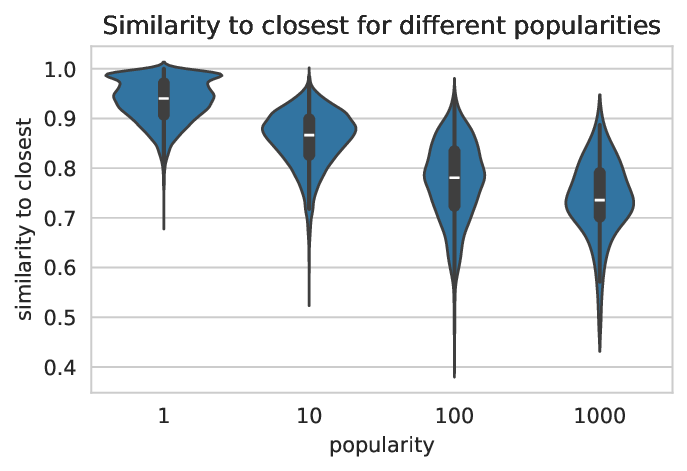}
    \caption{Similarity to closest item; histogram for all items and a violin plot (KDE on Y axis) for items grouped by popularity}
    \label{fig:sim}
\vspace{-.cm}
\end{wrapfigure}

Another symptom of bias is related to the distribution of item embeddings. We used the distance to the closest item as a measure of distinguishability. Figure \ref{fig:sim} presents the histogram of this measure for all items and a KDE plot with respect to the popularity of the items. We can see that the distribution is nonsymmetrical, with a long tail on smaller values, and two-modal, with a second modality for very close items. It also depends on the popularity of the item. More popular items are more distinguishable.

The biases learned by the model are then further propagated in the recommendations. We compared the SR-GNN model with two simple baselines: 
    recommending \textbf{random} items, with no information from the data (no bias),
     and recommending the most frequent successor (\textbf{bigram}), which we expect to capture the exposure related to the last item in the session. 
We took into account several criteria: the accuracy of recommendations measured with hit-rate, popularity bias measured with Average Recommendation Popularity (ARP), the width of new exposure measured with coverage and mimicking exposure from training dataset approximated with Intersection over Union (IoU) with respect to bigram recommendations. We also compare the results on both the training and testing data to capture how well the considered approaches generalize. All metrics were calculated for $K=20$ recommended items. 

\begin{table}[!ht]
    \centering
    \caption{Basic statistics of 20 recommendations on YooChoose 1/64 (21869 items), hit-rate, Average Recommendation Popularity (ARP), coverage, and Intersection over Union (IoU) with respect to bigram model recommendations.
    }
    \label{tab:preliminary}
    \begin{tabular}{|l|r|r|r|} \hline
        metric & random & bigram & SR-GNN \\ \hline
        hit-rate train &  0.0008 & 0.8293 & 0.7717 \\
        hit-rate test & 0.0009 & 0.5505 & 0.6301 \\
        ARP train & 79.2066 & 442.0736 & 540.8004 \\
        ARP test & 77.7978 & 445.9956 & 549.1695 \\
        coverage train & 1.0000 & 1.0000 & 0.8900 \\
        coverage test & 1.0000 & 1.0000 & 0.7700 \\
        IoU train & 0.0005 & -- & 0.3261 \\
        IoU test & 0.0005 & -- & 0.3216 \\ \hline
    \end{tabular}
\end{table}

The results can be found in Table \ref{tab:preliminary}. Recommendations based on bigrams are surprisingly accurate, although the approach is strongly overfitted. SR-GNN captures more complex relationships between items, making the recommendations more accurate and the model more general. The ARP on the test dataset for the bigram model is 5.7 times higher than for random recommendations, and for SR-GNN it is almost 7.1 times higher. It shows how strong the bias is in the data and that it is even enhanced by the SR-GNN model. IoU close to $\frac 1 3$ means that almost half of the SR-GNN and the bigram model recommendations overlap. It suggests that we tightly stick to the historical exposure.

\section{Proposed Approach} \label{sec:approach}





Data uncertainty, popularity bias, and exposure bias are closely related challenges in the development of SBRS. Their symptoms are observable both in the item embeddings and in the recommendations. We propose to address those challenges jointly using a mathematical formulation of user interest as a stochastic process in the latent space. A session is then considered as a realization of this stochastic process.
We propose a model-agnostic implementation of this approach consisting of three elements: improving the quality of item embeddings to mitigate popularity bias, modeling user interest as dense random variables to capture uncertainty, and introducing fake targets in the data to simulate extended exposure. 



\subsection{User Interest as a Stochastic Process} \label{sec:sp}



A stochastic process is a collection of random variables $\{X(t, \omega): t \in T\}$, where $T$ is the index set, and $\omega \in \Omega$ is an elementary event. For simplicity, let us denote $X(t, \omega)$ as $X_t$. If we observe a random process, we record its realization. 

We propose treating user interest at a given time stamp as an elementary event $\omega_{m, l}$ and considering the random variable $X_{m, l}$ valued in the latent space. This formulation allows us to capture the uncertainty of user behavior. In this view, user sessions present in the data are realizations of a stochastic process of user interest. 

In the recommendation task, we should rather recover the underlying user interest ($X_{m, l}$) than mimic the historical data ($s_{m, l}$). Let us note that a stochastic process can have multiple different realizations.



\subsection{Debiasing Item Representations}


Most SBRS store extracted information about the items in the embeddings. As we have shown in the preliminary studies, the popularity bias present in the data is encoded in the latent item representations. The identified syndromes of bias are encoding popularity in embedding norms \cite{Gupta2019NISERNI} and in the embedding distribution, as shown in the preliminary study.

To avoid these issues, we propose to use spherical embeddings (with an equal norm) and regularize the model for embedding uniformity using the Radial Basis Function (RBF) called the Gaussian potential:
\begin{equation} \label{eq:rbf}
    RBF_\tau(e_j, e_k) = e^{-\tau\Vert e_j - e_k \Vert_2^2},
    \mathcal L_{\text{unif}} = \log \mathbbm E [RBF_\tau(e_j, e_k)].
\end{equation}

Embeddings uniformly distributed on a sphere are also used in representation learning and are claimed to be high-quality representations \cite{pmlr-v119-wang20k, Yang_2023}. However, this technique has not been applied to SBRS so far.

\subsection{Dense User Interest}


Following the mathematical formulation presented in Section~\ref{sec:sp}, we propose to replace item embeddings with appropriate random variables when modeling session prefixes. 
The key aspect of implementing this approach is choosing a distribution that we assume is followed by latent representations of user interest. We propose using the von Mises-Fisher (vMF) distribution, which has a support on the sphere. The vMF distribution is parameterized with the mean direction $\mu$ and concentration parameter $\kappa$ and it is equivalent to the Gaussian distribution (with parameters $\mu$ and $\kappa^{-1}$) conditioned on the unit norm of the vector: 
\begin{equation}
   \mbox{vMF}(x, \mu, \kappa) \quad \sim \quad \mathcal N(x, \mu, \kappa ^{-1}) \quad | \quad \Vert x \Vert_2 = 1.
\end{equation}

The vMF distribution is also related to the way SBRS score items for a given session. As we describe in Section \ref{sec:emb}, having obtained an embedding of a session, which we can think of as predicted user interest, we calculate its dot product with item embeddings. If we want to consider probabilities of interest, we would transform the scores using the softmax function. Similarly, the PDF of the vMF distribution is proportional to the exponent of the dot product of a given point $x$ and the direction parameter $\mu$: $\mbox{vMF}(x, \mu, \kappa) \propto \exp \{\kappa x^T \mu\}$.


We model user interest varying from session to session, as those are anonymous and independent, and from timestamp to timestamp, as it may change over time. Thus, we only have one observation $s_{m, l}$ to estimate the distribution of user interest $X_{m, l}$. We use the unbiased estimator of the direction parameter $e_{m, l}$ and a fixed $\kappa$ (to avoid a degenerate distribution): \linebreak $X_{m,l} \sim \mbox{vMF}(x, e_{m, l}, \kappa)$.


To infer from dense user interest, we replace the representation of a session using embeddings $(e_{m, 1}, e_{m, 2}, \ldots, e_{m, L_m})$ with one using random variables $(X_{m, 1}, X_{m, 2}, \ldots, X_{m, L_m})$ and feed the model with a different realization of the stochastic process in each epoch. This allows the data uncertainty to be incorporated into the training procedure. 

\subsection{Extended Exposure}


\begin{wrapfigure}{r}{0.45\linewidth}
\vspace{-.5cm}
    \begin{center}
    \includegraphics[width=\linewidth]{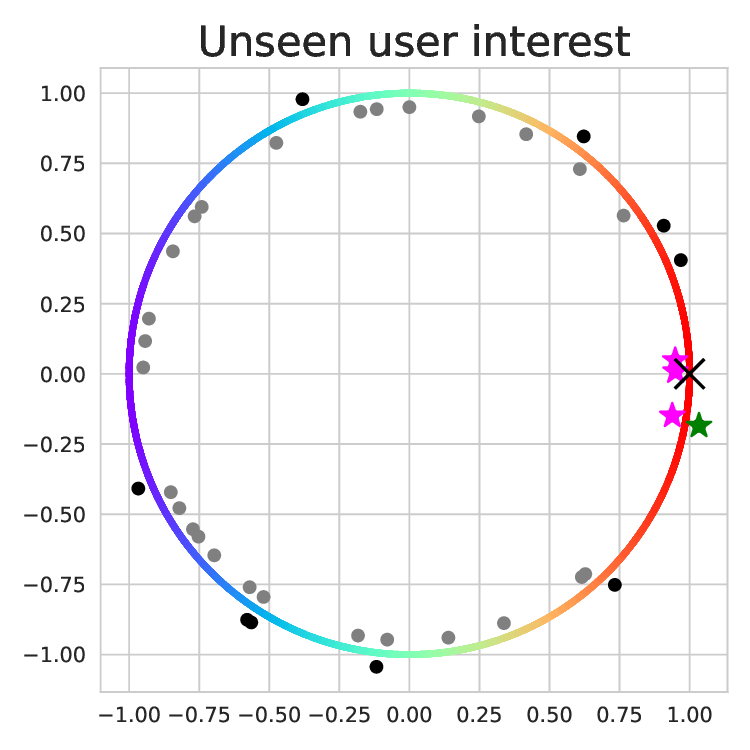}
    \caption{The circle is the dense spherical space of possible user interest. The latent user interest is marked with 'x'. The colours of the circle correspond to the alignment with user interest. The dots and stars are the representations of all considered items. The user has been exposed to items shifted outside the circle, those shifted inside remain unseen. The green star is the target present in the data. The magenta stars are possible, unseen targets, which we would like to discover by sampling \textit{fake targets}}
    \label{fig:fakes}    
    \end{center}
\end{wrapfigure}



As we formulate user interest as a stochastic process, we can also add that it depends on the exposure, ie. the distribution is strongly shifted towards visible items. Modeling it in a dense manner weakens this dependency. However, it is not applicable to the target items, as most SBRS are trained with the cross-entropy loss function. Thus, instead of sampling arbitrary points from the latent space, we will use a similar procedure to sample items, which the user might have also liked, and feed them into the loss function as \textit{fake targets}. It gives the effect of extended exposure, as the model is given that the user has seen additional items. The idea is presented in Figure \ref{fig:fakes}.




Fake target sampling is performed using the following procedure. First, we select an uncertainty parameter $\alpha$ describing, how much probability we want to redistribute from the original target.
Secondly, we select a minimal similarity parameter $\beta$ limiting the cosine similarity between the true and fake targets. From the targets meeting the similarity criterion, we sample a set $F_{m} = \{f_{m,1}, f_{m, 2}, \ldots, f_{m,P}\}$ of $P$ item IDs with probability proportional to vMF PDF with concentration parameter~$\kappa$.

If no item satisfies the similarity criterion, the probability of true targets remains equal to one. Introducing fake targets implies a modification of the loss function $\mathcal L_{rec}$. As SR-GNN is optimized with cross-entropy loss, the function is optimized with modified target probabilities:
\begin{equation}
    y_j^* = \begin{cases}
        \frac { \exp\{\kappa e_j^T e_{m,L_m}\}} {\sum_{p=1}^P  \exp\{\kappa e_{f_{m, l, p}}^T e_{m,L_m}\}} \cdot \alpha & \text{ if } j \in F_{m} \text{ (fake)}\\
         1 -  \sum_{p=1}^P y^*_{f_{m, p}} & \text{ if } s_{m,L_m} = i_j \text{ (true)}\\
        0 & \text{otherwise}
    \end{cases}
\end{equation}
and the recommendation loss formula becomes: $\mathcal L_{rec}^* = \sum_{j=1}^N y_j^* \cdot \log(\hat y_j)$,
where $\hat y$ is a soft-max normalized vector of dot products between the item embeddings and the predicted interest direction. Applying this loss function, we try to march the predicted user interest, the true target and the fake targets with some weights (given by $y^*$). The final loss is given by $\mathcal L = \mathcal L_{rec}^* + \lambda \mathcal L_{\text{unif}}$.

\section{Experiments}\label{sec:experiments}



We have proposed a model-agnostic component allowing for hybrid deep-stochastic modeling of SBRS. In this section, we report on computational experiments conducted to investigate whether this helps mitigate the problems presented in the preliminary study. We aim to answer the following research questions:

\begin{itemize}
    \item \textbf{RQ1:} Can the stochastic component expand the exposure of the test user group?
    \item \textbf{RQ2:} Can the stochastic component mitigate the popularity bias in the model and in the recommendations? 
    \item \textbf{RQ3:} Can the stochastic component improve the generalization of the SBRS model?
\end{itemize}

The presented approach allows to address uncertainty, popularity and exposure bias jointly. In the experiments, we focus on measurable symptoms of bias.




\subsection{Setup}


\noindent \textbf{Datasets.}
In order to validate our approach, we performed computational experiments on real-world datasets, YooChoose and Diginetica, which are popular benchmarks for SBRS with varying levels of bias. 

We apply the standard preprocessing method 
\cite{Yu_2020, Wu2019ke} to both datasets, starting with removing sessions of length 1 and items appearing less than 5 times, then extracting the test data (sessions of the last day) and the train data (the remaining sessions). For both the test and the train dataset, similar to \cite{Wu2019ke}, we applied the sequence splitting process, that is, each session $(s_{m, 1}, s_{m, 2}, \ldots, s_{m, L_m})$ gave an input sequence $(s_{m, 1}, s_{m, 2}, \ldots, s_{m, k-1})$ and a target $s_{m, k}$, for $k=2, 3, \ldots, L_m$. 
 From this, we created the \textbf{Diginetica} dataset. We kept 1/64 of the YooChoose train sessions, as suggested in \cite{improved_rnn_for_session_based_recommendations} and used in \cite{Yu_2020, Wu2019ke} to create \textbf{YC~1/64}.  Afterwards, in all the cases, the items not appearing in the train data were removed from the test data.

\begin{table}[bp]
\caption{Statistics of the datasets. 
}
\label{tab:datasets}
\begin{center}
    \begin{tabular}{|l|r|r|r|} \hline
         dataset & train items & train sessions & test session \\ \hline
        Diginetica & 42999 & 727276 & 60632 \\
        YC 1/64 & 16191 & 368626 & 55239 \\ \hline
         001~-~YC~1/64  & 14531 & 289871 & 43820 \\
         002~-~YC~1/64  & 13299 & 224721 & 40236 \\
         003~-~YC~1/64  & 12303 & 182022 & 30699 \\
         004~-~YC~1/64  & 11464 & 142385 & 23555 \\
         005~-~YC~1/64  & 6529 & 17230 & 4982 \\ \hline
    \end{tabular}
\end{center}
\end{table}
 
 Furthermore, to study the influence of the balance of rare-popular items, we provided modified versions of the YooChoose dataset, similarly to the methodology used in \cite{GraphSMOTE}. 
 In order to make the data unbalanced, before the above preprocessing procedure and extracting the train data,  $20\%, 40\%, 60\%, 80\% \text{ or } 100\%$ of the most popular items (randomly chosen from the first quartile of popularity) have been removed from the data
 to create the datasets \textbf{001~-~YC~1/64},  \textbf{002~-~YC~1/64},  \textbf{003~-~YC~1/64},  \textbf{004~-~YC~1/64},  \textbf{005~-~YC~1/64}, respectively. 
 It is worth noticing that popular items strongly dominate in the data, so removing them causes significant decreases in the number of sessions, because many sessions became empty or of length $1$ (impossible to split for a nonempty prefix and target), and consequently, removing the new sessions of length $1$ causes removing some unpopular products. 
  Table \ref{tab:datasets} presents statistics for the data sets.


\begin{table*}[b]
    \centering
    \caption{Overall results on standard datasets. The best results for all the models have been \underline{underlined}, best results for the graph-based models (NISER, SR-GNN, TAGNN, noisy -- our approach) have been \textbf{bold}.}
    \label{tab:overall}
    \begin{tabular}{|l|r|r|r|r|r|r|r|} \hline 
        model &  Hit-Rate Train & Hit-Rate Test & Coverage Train & Coverage Test & ARP Train & ARP Test & RBF \\ \hline
        \multicolumn{8}{c}{Diginetica} \\ \hline
        random & 0.0005 & 0.0004 & \underline{100.0000} & \underline{100.0000} & \underline{16.9196} & \underline{16.8887} & - \\
        bigram & \underline{0.8516} & 0.3480 & 100.0000 & 99.9977 & 55.1153 & 53.5065  & - \\ \hline
        NISER & 0.7369 & 0.4831 & 98.7302 & 91.0161 & 61.9805 & \textbf{59.7067} & \underline{\textbf{0.0606}} \\
        SR-GNN & \textbf{0.7794} & 0.4656  & 98.1883 & 91.9719 & \textbf{61.3056} & 60.8660 & 0.0633 \\
        TAGNN & 0.7399 & 0.4869 & 98.7744 & 90.7533 & 62.7503 & 61.0402 & 0.0642 \\ \hline
        noisy & 0.7448 & \underline{\textbf{0.5109}} & \textbf{99.9372} & \textbf{93.2650} & 64.8328 & 64.4680 & 0.0632 \\ \hline
        \multicolumn{8}{c}{YC 1/64} \\ \hline
        random & 0.0013 & 0.0011 & \underline{100.0000} & \underline{100.0000} & \underline{22.7245} & \underline{22.7522} & - \\
        bigram & \underline{0.8214} & 0.6496 & \underline{100.0000} & 99.9815  & 541.5820 & 422.3570 & - \\ \hline
        NISER & 0.7839 & \underline{\textbf{0.7154}}  & 90.5194 & 68.9828 & 577.5689 & 452.8599 & 0.0630 \\
        SR-GNN & \textbf{0.7922} & 0.7020  & 92.9343 & 71.8053 & 570.2716 & 440.6263 & 0.0584 \\
        TAGNN & 0.7848 & 0.7142  & 90.6800 & 68.9828 & 573.5134 & 450.1954 & 0.0654 \\ \hline
        noisy & 0.7809 & 0.7083  & \textbf{99.6418} & \textbf{76.0917} & \textbf{548.8170} & \textbf{432.2718} & \underline{\textbf{0.0551}} \\ \hline
    \end{tabular}
\end{table*}

\medskip
\noindent \textbf{Baselines.}
In computational experiments, we compare our results to 2 naive approaches  described in Section \ref{sec:preliminaries} (random and bigram model) and 3 popular SBRS based on graph neural networks, SR-GNN \cite{Wu2019ke}, TAGNN \cite{Yu_2020}, and NISER \cite{Gupta2019NISERNI}.


\textbf{SR-GNN} \cite{Wu2019ke} is described in detail in Section \ref{sec:background}.
\textbf{NISER} \cite{Gupta2019NISERNI} extends SR-GNN by addressing the popularity bias problem by introducing a mechanism of normalizing representations. \textbf{TAGNN} \cite{Yu_2020} extends SR-GNN with an additional target attention mechanism, where the secondary item embeddings, evaluated by the GGNN layer, are further processed using a target attention vector.

\begin{table*}[!ht] 
    \centering
    \caption{Overall results on YC 1/64 modifications.
    Best results for each dataset have been \underline{underlined}.}
    \label{tab:modified-YC}
   \begin{tabular}{|l|l|r|r|r|r|r|r|r|} \hline
        dataset & model &  Hit-Rate Train & Hit-Rate Test & Coverage Train & Coverage Test & ARP Train & ARP Test & RBF \\ \hline
        \multirow[c]{4}{*}{001} & NISER  & 0.7931 & 0.7161 & 93.3728 & 69.4859 & 507.2249 & 375.9715 & 0.0633 \\
         & SR-GNN & \underline{0.8029} & 0.7055 & 95.6232 & 71.6744 & 501.7716 & 368.0752 & 0.0584 \\
         & TAGNN & 0.7948 & \underline{0.7174} & 93.6618 & 69.9608 & 506.6953 & 371.1567 & 0.0622 \\ 
         & noisy  & 0.7923 & 0.7140 & \underline{99.7591} & \underline{76.2439} & \underline{479.1834} & \underline{353.9358} & \underline{0.0522} \\ \hline
        \multirow[c]{4}{*}{002} & NISER & 0.7995 & \underline{0.7274} & 95.2478 & 71.6670 & 443.3394 & 367.6830 & 0.0596 \\
         & SR-GNN & \underline{0.8083} & 0.7160 & 96.9171 & 72.3438 & 442.8813 & 362.1181 & 0.0584 \\
         & TAGNN & 0.7994 & \underline{0.7274} & 95.0297 & 71.2685 & 438.5160 & 365.4440 & 0.0561 \\
         & noisy &  0.7975 & 0.7225 & \underline{99.7368} & \underline{76.2087} & \underline{417.8626} & \underline{349.0806} & \underline{0.0531} \\ \hline
        \multirow[c]{4}{*}{003} & NISER  & 0.8244 & \underline{0.7315} & 95.6840 & 69.5603 & 380.1184 & 321.6723 & 0.0609 \\
         & SR-GNN & \underline{0.8321} & 0.7237 & 97.3828 & 71.1290 & 377.3208 & 317.0184 & 0.0583 \\
         & TAGNN & 0.8249 & 0.7313 & 95.7571 & 70.3975 & 378.3776 & 320.0574 & 0.0604 \\
         & noisy & 0.8226 & 0.7285 & \underline{99.4798} & \underline{75.0792} & \underline{363.6459} & \underline{307.6846} & \underline{0.0546} \\ \hline
        \multirow[c]{4}{*}{004} & NISER & 0.8257 & \underline{0.7256} & 95.8043 & 68.4054 & 346.6219 & 242.4871 & 0.0612 \\
         & SR-GNN & \underline{0.8411} & 0.7232 & 97.6971 & 69.3214 & 343.6278 & 241.2055 & 0.0588 \\
         & TAGNN & 0.8306 & 0.7255 & 96.2055 & 69.7052 & 343.8487 & 244.2331 & 0.0625 \\
         & noisy & 0.8293 & 0.7255 & \underline{99.3545} & \underline{74.7819} & \underline{329.8606} & \underline{230.7884} & \underline{0.0563} \\ \hline
        \multirow[c]{4}{*}{005} & NISER & 0.8337 & 0.7051 & 86.1694 & 52.2745 & \underline{15.1847} & 37.0339 & 0.0576 \\
         & SR-GNN & \underline{0.9481} & 0.7318 & 96.2169 & 54.3728 & 15.5628 & 36.9127 & 0.0594 \\
         & TAGNN & 0.8328 & 0.7130 & 87.0271 & 49.7779 & 15.2988 & \underline{36.7836} & \underline{0.0573} \\
         & noisy & 0.9353 & \underline{0.7336} & \underline{96.7529} & \underline{61.0354} & 16.1653 & 37.6747 & 0.0603 \\ \hline
    \end{tabular}
\end{table*}
 
\medskip
\noindent \textbf{Metrics.}
We evaluate the models trained with the following metrics:
hit-rate,
coverage, 
Average Recommendation Popularity (ARP),
representation ratio,
Radial Basis Function (RBF),
cosine similarity with the closest item.
The models are evaluated with respect to the targets originally present in the data. We used $K=20$ recommendations.

\noindent \textbf{Hyper-parameters.} We trained all the models using the Adam optimizer with a learning rate of $0.001$, batch size $32$, and learning rate decay $0.1$ after each epoch. Our approach was trained with $\kappa = 250$, $\lambda=0.5$, $F=10$, and $\alpha=10\%$ target uncertainty with threshold $\beta=0$ and all other hyper-parameters as above.
We trained all the models from scratch with $15$ epochs.

\subsection{Results}


Table \ref{tab:overall} presents the results obtained for the Diginetica and YC 1/64 datasets. 
Recommending random items does not utilize any information present in the data, and thus the method is inaccurate and unbiased. The simple bigram model overfits the training data. Although it struggles to generalize, its accuracy is quite good and coverage is almost perfect. However, we can see that it is much more prone to recommend more popular items. The ARP ratio (with respect to random recommendations) on the test set for Diginetica and YC 1/64 are approximately $3.17$ and $18.56$ (please note the different preprocessing then in preliminary research). The results obtained suggest the differences between these two datasets: Diginetica dataset is less biased with popularity and less trivial to model user interest than the YC 1/64 dataset. 

When comparing the GNN-based approaches, we can see that the enhancement of SRGNN with the stochastic component improves coverage for both datasets, which extends the exposure related to recommending items to new users. For Diginetica, which is less biased with popularity, our approach increased hit-rate for test data and, therefore, improved the model generalization. However, it increased the popularity of recommended items. The opposite effect has been achieved for the YC 1/64 dataset, which is far more biased with popularity. The accuracy of recommendations was comparable to that of the backbone model and did not outperform other baselines; however, the popularity of the recommendations decreased. The proposed approach improved the performance of the model in the criteria of the highest bias for each dataset. 

Table \ref{tab:modified-YC} presents the results for the modified versions of the YC 1/64 dataset enabling to better understand how popularity bias affects the performance of the models. The differences between the datasets containing very popular items (001, 002, 003, 004) and one with all popular items excluded (005) mimic the differences between the standard YC 1/64 and Diginetica datasets. For datasets with strong popularity bias, the Hit-Rate values are comparable, and we got an improvement in terms of coverage, recommendation popularity, and item uniformity. For the debiased dataset, we got a slightly higher Hit-Rate and considerably higher coverage, with slightly worse ARP and RBF. What is interesting, for the first 4 modifications, all models performed quite similar, while for the most strongly modified dataset, hit-rate, and coverage are visibly lower for the more complicated approaches NISER and TAGNN.

\begin{table}[!bp]
    \centering
    \caption{Representation of items from popularity quartiles (Q1 - most, Q4 - least popular). For items below the popularity median (Q3, Q4),  we \underline{underlined} the highest representation values.}
    \label{tab:representation}
    \begin{tabular}{|l|l|r|r|r|r|} \hline
         data set & model & Q1 & Q2 & Q3 & Q4 \\ \hline
        \multirow[c]{4}{*}{001} & NISER & 0.8417 & 0.1411 & 0.0160 & 0.0012 \\
         & SR-GNN & 0.8384 & 0.1424 & 0.0179 & \underline{0.0014} \\
         & TAGNN & 0.8382 & 0.1433 & 0.0173 & 0.0012 \\
         & noisy & 0.8337 & 0.1459 & \underline{0.0190} & \underline{0.0014} \\  \hline
        \multirow[c]{4}{*}{002} & NISER & 0.8263 & 0.1540 & 0.0180 & 0.0017 \\
         & SR-GNN & 0.8216 & 0.1559 & 0.0210 & 0.0015 \\
         & TAGNN & 0.8248 & 0.1545 & 0.0191 & 0.0015 \\
         & noisy & 0.8187 & 0.1583 & \underline{0.0212} & \underline{0.0018} \\ \hline
        \multirow[c]{4}{*}{003} & NISER & 0.7747 & 0.1974 & 0.0259 & 0.0020 \\
         & SR-GNN & 0.7703 & 0.1991 & 0.0284 & {0.0022} \\
         & TAGNN & 0.7730 & 0.1991 & 0.0258 & 0.0021 \\
         & noisy & 0.7665 & 0.2020 & \underline{0.0293} & \underline{0.0022} \\ \hline
        \multirow[c]{4}{*}{004} & NISER & 0.7071 & 0.2568 & 0.0332 & \underline{0.0029} \\
         & SR-GNN & 0.6994 & 0.2605 & \underline{0.0374} & 0.0028 \\
         & TAGNN & 0.7043 & 0.2587 & 0.0343 & 0.0027 \\
         & noisy & 0.6965 & 0.2634 & 0.0373 & 0.0028 \\ \hline
        \multirow[c]{4}{*}{005} & NISER &  - & 0.8620 & 0.1273 & 0.0107 \\
         & SR-GNN & - & 0.8539 & 0.1354 & 0.0107 \\
         & TAGNN & - & 0.8626 & 0.1274 & 0.0100 \\
         & noisy & - & 0.8520 & \underline{0.1365} & \underline{0.0115} \\ \hline
    \end{tabular}
\end{table}

Every of the considered models (NISER, SRGNN, TAGNN, our noisy approach) has the best Hit-Rate for at least one of the datasets. In addition, the results for the datasets with custom preprocessing from the preliminary studies in Section \ref{sec:preliminaries} and the preprocessing well-established in the literature vary a lot. When we exclude selected items and sessions from the data in the preprocessing, we manipulate the level of difficulty of the recommendation task. This makes the results for different datasets (and their preprocessed versions) incomparable. Having the perspective of multiple data preprocessing scenarios, we can see that all the graph-based baselines give results which are generally comparable.

To better understand the popularity bias in the models, we also evaluated the symptoms of popularity bias detected in the preliminary studies. Table \ref{tab:representation} presents how often items from different popularity quartiles are recommended. Although we provide improvement for almost all datasets (excluding 004), the items with popularity below the median are recommended very rarely. 
However, we have made considerable progress when it comes to embedding quality. 

\begin{figure}[!ht]
    \centering
    \includegraphics[width=1\linewidth]{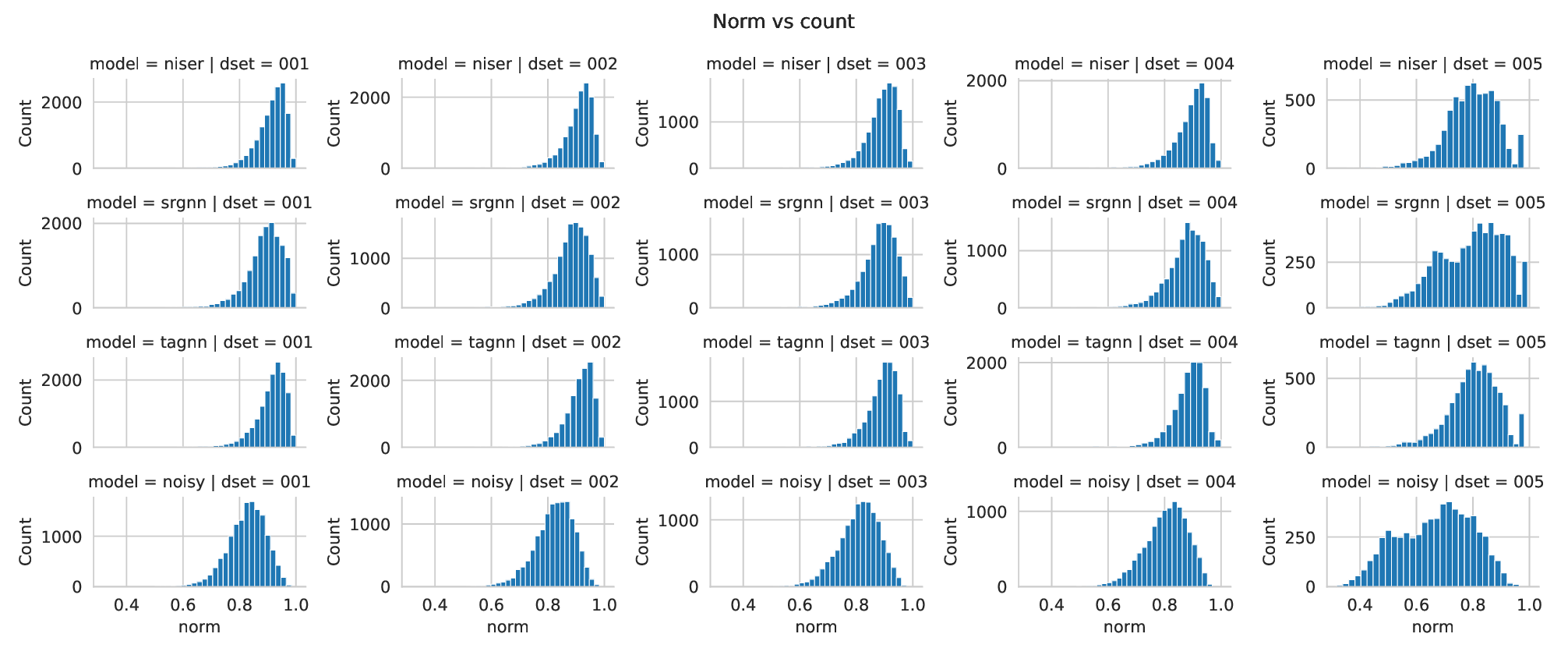}
    \caption{Histogram of similarity with the closest item (calculated with cosine similarity between item embeddings). Each column corresponds to a~model (noisy SR-GNN, NISER, SR-GNN, TAGNN) and each row to a~dataset
    }
    \label{fig:cosine}
\end{figure}

Figure \ref{fig:cosine} presents the distribution of cosine similarity to the closest neighbor for all items. Our approach gives a visible shift towards lower values, which means, it makes the items more distinguishable. Thus, we reduce the symptoms of popularity bias. 

\section{Conclusions} \label{sec:conclusions}

In this paper, we considered the issues of data uncertainty, popularity bias, and exposure bias in SBRS, which are both important research topics and  difficulties in practical applications. We discussed how these challenges interconnect and studied several measurable symptoms of bias in item embeddings and recommendations. To address those, we proposed a formulation of user interest as a stochastic process in the latent space together with an implementation of this mathematical concept. The proposed approach consists of three elements: debiasing item representations, modeling dense user interest, and extending historical user exposure in training. It can be incorporated into a neural network-based model architecture, resulting in a hybrid deep-stochastic approach to modeling user sessions. 

We conducted computational experiments to validate our approach implemented on the SR-GNN backbone. The broad evaluation showed that the presented stochastic component allows for mitigating exposure and popularity bias, as well as improves generalization of the model. The performance of our approach depended on the characteristics of the data, the key advancements were made in the aspects that represent the most challenging aspects of a given dataset.

\bibliographystyle{IEEEtran}
\bibliography{bibf}

\end{document}